%% file: arxiv.tex
\documentclass[final]{cvpr}
\usepackage{times}
\usepackage{epsfig}
\usepackage{graphicx}
\usepackage{amsmath}
\usepackage{amssymb}

\usepackage[pagebackref=true,breaklinks=true,colorlinks,bookmarks=false]{hyperref}
\usepackage[table]{xcolor}
\usepackage[shortlabels]{enumitem}
\usepackage{hyperref}
\usepackage{booktabs}
\usepackage{pifont}
\usepackage{multirow}
\usepackage{floatrow}
\newfloatcommand{capbtabbox}{table}[][\FBwidth]

\makeatletter
\@ifundefined{c@rownum}{%
  \let\c@rownum\rownum
}{}
\@ifundefined{therownum}{%
  \def\therownum{\@arabic\rownum}%
}{}
\makeatother

\DeclareFontFamily{U}{mathx}{\hyphenchar\font45}
\DeclareFontShape{U}{mathx}{m}{n}{
      <5> <6> <7> <8> <9> <10>
      <10.95> <12> <14.4> <17.28> <20.74> <24.88>
      mathx10
      }{}
\DeclareSymbolFont{mathx}{U}{mathx}{m}{n}
\DeclareFontSubstitution{U}{mathx}{m}{n}
\DeclareMathAccent{\widecheck}{0}{mathx}{"71}
\DeclareMathOperator*{\argmax}{arg\,max}

\newcommand{\cmark}{\ding{51}}
\newcommand{\xmark}{\ding{55}}
\newcommand{\bs}[1]{\boldsymbol{#1}}
\definecolor{lightgray}{gray}{0.9}

\def\cvprPaperID{457} %
\def\confYear{CVPR 2021}

\begin{document}

\title{SimPoE: Simulated Character Control for 3D Human Pose Estimation}

\author{
Ye Yuan\textsuperscript{1} \qquad Shih-En Wei\textsuperscript{2} \qquad Tomas Simon\textsuperscript{2} \qquad Kris Kitani\textsuperscript{1} \qquad Jason Saragih\textsuperscript{2} \\[1mm]
\textsuperscript{1}Carnegie Mellon University \qquad \textsuperscript{2}Facebook Reality Labs \\[1mm]
{ \url{https://www.ye-yuan.com/simpoe}} \\
}
\maketitle

\begin{abstract}
\vspace{-3mm}
Accurate estimation of 3D human motion from monocular video requires modeling both kinematics (body motion without physical forces) and dynamics (motion with physical forces). To demonstrate this, we present \mbox{SimPoE}, a \textbf{Sim}ulation-based approach for 3D human \textbf{Po}se \textbf{E}stimation, which integrates image-based kinematic inference and physics-based dynamics modeling. SimPoE learns a policy that takes as input the current-frame pose estimate and the next image frame to control a physically-simulated character to output the next-frame pose estimate. The policy contains a learnable kinematic pose refinement unit that uses 2D keypoints to iteratively refine its kinematic pose estimate of the next frame. Based on this refined kinematic pose, the policy learns to compute dynamics-based control (e.g., joint torques) of the character to advance the current-frame pose estimate to the pose estimate of the next frame. This design couples the kinematic pose refinement unit with the dynamics-based control generation unit, which are learned jointly with reinforcement learning to achieve accurate and physically-plausible pose estimation. Furthermore, we propose a meta-control mechanism that dynamically adjusts the character's dynamics parameters based on the character state to attain more accurate pose estimates. Experiments on large-scale motion datasets demonstrate that our approach establishes the new state of the art in pose accuracy while ensuring physical plausibility.
\end{abstract}

\vspace{-5mm}
\section{Introduction}
\vspace{-2mm}
\label{sce:intro}
We aim to show that accurate 3D human pose estimation from monocular video requires modeling both kinematics and dynamics. Human dynamics, \emph{i.e.,} body motion modeling with physical forces, has gained relatively little attention in 3D human pose estimation compared to its counterpart, kinematics, which models motion without physical forces. There are two main reasons for the disparity between these two equally important approaches. First, kinematics is a more direct approach that focuses on the geometric relationships of 3D poses and 2D images; it sidesteps the challenging problem of modeling the physical forces underlying human motion, which requires significant domain knowledge about physics and control. Second, compared to kinematic measurements such as 3D joint positions, physical forces present unique challenges in their measurement and annotation, which renders standard supervised learning paradigms unsuitable. Thus, almost all state-of-the-art methods~\cite{pavlakos2019expressive,xiang2019monocular,kolotouros2019learning,kocabas2020vibe,moon2020i2l} for 3D human pose estimation from monocular video are based only on kinematics.
Although these kinematic methods can estimate human motion with high pose accuracy, they often fail to produce physically-plausible motion. Without modeling the physics of human dynamics, kinematic methods have no notion of force, mass or contact; they also do not have the ability to impose physical constraints such as joint torque limits or friction. As a result, kinematic methods often generate physically-implausible motions with pronounced artifacts: body parts (\emph{e.g.,} feet) penetrate the ground; the estimated poses are jittery and vibrate excessively; the feet slide back and forth when they should be in static contact with the ground. All these physical artifacts significantly limit the application of kinematic pose estimation methods. For instance, jittery motions can be misleading for medical monitoring and sports training; physical artifacts also prevent applications in computer animation and virtual/augmented reality since people are exceptionally good at discerning even the slightest clue of physical inaccuracy~\cite{reitsma2003perceptual,hoyet2012push}.

To improve the physical plausibility of estimated human motion from video, recent work~\cite{li2019estimating,rempe2020,shimada2020physcap} has started to adopt the use of dynamics in their formulation. These methods first estimate kinematic motion and then use physics-based trajectory optimization to optimize the forces to induce the kinematic motion. Although they can generate physically-grounded motion, there are several drawbacks of trajectory optimization-based approaches. First, trajectory optimization entails solving a highly-complex optimization problem at test time. This can be computationally intensive and requires the batch processing of a temporal window or even the entire motion sequence, causing high latency in pose predictions and making it unsuitable for interactive real-time applications. Second, trajectory optimization requires simple and differentiable physics models to make optimization tractable, which can lead to high approximation errors compared to advanced and non-differentiable physics simulators (\emph{e.g.,} MuJoCo~\cite{todorov2012mujoco}, Bullet~\cite{coumans2010bullet}). Finally and most importantly, the application of physics in trajectory optimization-based methods is implemented as a post-processing step that projects a given kinematic motion to a physically-plausible one. Since it is optimization-based, there is no learning mechanism in place that tries to match the optimized motion to the ground truth. As such, the resulting motion from trajectory optimization can be physically-plausible but still far from the ground-truth, especially when the input kinematic motion is inaccurate.

To address these limitations, we present a new approach, SimPoE (\emph{\textbf{Sim}ulated Character Control for Human \textbf{Po}se \textbf{E}stimation}), that tightly integrates image-based kinematic inference and physics-based dynamics modeling into a joint learning framework. Unlike trajectory optimization, SimPoE is a causal temporal model with an integrated physics simulator. Specifically, SimPoE learns a policy that takes the current pose and the next image frame as input, and produces controls for a proxy character inside the simulator that outputs the pose estimate for the next frame. To perform kinematic inference, the policy contains a learnable kinematic pose refinement unit that uses image evidence (2D keypoints) to iteratively refine a kinematic pose estimate. Concretely, the refinement unit takes as input the gradient of keypoint reprojection loss, which encodes rich information about the geometry of pose and keypoints, and outputs the kinematic pose update. Based on this refined kinematic pose, the policy then computes a character control action, \emph{e.g.,} target joint angles for the character's proportional-derivative (PD) controllers, to advance the character state and obtain the next-frame pose estimate. This policy design couples the kinematic pose refinement unit with the dynamics-based control generation unit, which are learned jointly with reinforcement learning (RL) to ensure both accurate and physically-plausible pose estimation. At each time step, a reward is assigned based on the similarity between the estimated motion and the ground truth. To further improve pose estimation accuracy, SimPoE also includes a new control mechanism called meta-PD control. PD controllers are widely used in prior work~\cite{peng2017learning,peng2018deepmimic,yuan2019ego} to convert the action produced by the policy into the joint torques that control the character. However, the PD controller parameters typically have fixed values that require manual tuning, which can produce sub-optimal results. Instead, in meta-PD control, SimPoE's policy is also trained to dynamically adjust the PD controller parameters across simulation steps based on the state of the character to achieve a finer level of control over the character's motion. 

We validate our approach, SimPoE, on two large-scale datasets, Human3.6M~\cite{ionescu2013human3} and an in-house human motion dataset that also contains \emph{detailed finger motion}. We compare \mbox{SimPoE} against state-of-the-art monocular 3D human pose estimation methods including both kinematic and physics-based approaches. On both datasets, SimPoE outperforms previous art in both pose-based and physics-based metrics, with significant pose accuracy improvement over prior physics-based methods. We further conduct extensive ablation studies to investigate the contribution of our proposed components including the kinematic refinement unit, meta-PD control, as well as other design choices.

The main contributions of this paper are as follows: (1)~We present a joint learning framework that tightly integrates image-based kinematic inference and physics-based dynamics modeling to achieve accurate and physically-plausible 3D human pose estimation from monocular video. (2)~Our approach is causal, runs in real-time without batch trajectory optimization, and addresses several drawbacks of prior physics-based methods. (3)~Our proposed meta-PD control mechanism eliminates manual dynamics parameter tuning and enables finer character control to improve pose accuracy. (4) Our approach outperforms previous art in both pose accuracy and physical plausibility. (5) We perform extensive ablations to validate the proposed components to establish good practices for RL-based human pose estimation.

\section{Related Work}
\noindent\textbf{Kinematic 3D Human Pose Estimation.}
Numerous prior works estimate 3D human joint locations from monocular video using either two-stage~\cite{dabral2018learning,rayat2018exploiting,pavllo20193d} or end-to-end~\cite{mehta2017vnect,mehta2018single} frameworks. On the other hand, parametric human body models~\cite{anguelov2005scape,loper2015smpl,pavlakos2019expressive} are widely used as the human pose representation since they additionally provide skeletal joint angles and a 3D body mesh. Optimization-based methods have been used to fit the SMPL body model~\cite{loper2015smpl} to 2D keypoints extracted from an image ~\cite{bogo2016keep,lassner2017unite}. Alternatively, regression-based approaches use deep neural networks to directly regress the parameters of the SMPL model from an image~\cite{tung2017self,tan2017indirect,pavlakos2018learning,omran2018neural,kanazawa2018end,guler2019holopose}, using weak supervision from 2D keypoints~\cite{tung2017self,tan2017indirect,kanazawa2018end} or body part segmentation~\cite{omran2018neural,pavlakos2018learning}. Song \emph{et al.}~\cite{song2020human} propose neural gradient descent to fit the SMPL model using 2D keypoints. Regression-based~\cite{kanazawa2018end} and optimization-based~\cite{bogo2016keep} methods have also been combined to produce pseudo ground truth from weakly-labeled images~\cite{kolotouros2019learning} to facilitate learning. Recent work~\cite{arnab2019exploiting,huang2017towards,kanazawa2019learning,sun2019human,kocabas2020vibe,luo20203d} starts to exploit the temporal structure of human motion to estimate smooth motion. Kanazawa \emph{et al.}~\cite{kanazawa2019learning} model human kinematics by predicting past and future poses. Transformers~\cite{vaswani2017attention} have also been used to improve the temporal modeling of human motion~\cite{sun2019human}. All the aforementioned methods disregard human dynamics, \emph{i.e.,} the physical forces that generate human motion. As a result, these methods often produce physically-implausible motions with pronounced physical artifacts such as jitter, foot sliding, and ground penetration.

\vspace{1mm}
\noindent\textbf{Physics-Based Human Pose Estimation.} A number of works have addressed human dynamics for 3D human pose estimation. Most prior works~\cite{brubaker2009estimating,wei2010videomocap,vondrak2012video,zell2017joint,yuan2019ego,rempe2020,shimada2020physcap} use trajectory optimization to optimize the physical forces to induce the human motion in a video. As discussed in Sec.~\ref{sce:intro}, trajectory optimization is a batch procedure which has high latency and is typically computationally expensive, making it unsuitable for real-time applications. Furthermore, these methods cannot utilize advanced physics simulators with non-differentiable dynamics. Most importantly, there is no learning mechanism in trajectory optimization-based methods that tries to match the optimized motion to the ground truth. Our approach addresses these drawbacks with a framework that integrates kinematic inference with RL-based character control, which runs in real-time, is compatible with advanced physics simulators, and has learning mechanisms that aim to match the output motion to the ground truth. Although prior work~\cite{yuan20183d,yuan2019ego,isogawa2020optical} has used RL to produce simple human locomotions from videos, these methods only learn policies that coarsely mimic limited types of motion instead of precisely tracking the motion presented in the video. In contrast, our approach can achieve accurate pose estimation by integrating images-based kinematic inference and RL-based character control with the proposed policy design and meta-PD control.

\vspace{1mm}
\noindent\textbf{Reinforcement Learning for Character Control.}
Deep RL has become the preferred approach for learning character control policies with manually-designed rewards~\cite{liu2017learning,liu2018learning,peng2018deepmimic,peng2018sfv}. GAIL~\cite{ho2016generative} based methods are proposed to learn character control without reward engineering~\cite{merel2017learning,wang2017robust}. To produce long-term behaviors, prior work has used hierarchical RL to control characters to achieve high-level tasks~\cite{merel2018neural,merel2018hierarchical,peng2019mcp,merel2020catch}. Recent work also uses deep RL to learn user-controllable policies from motion capture data for character animation~\cite{bergamin2019drecon,park2019learning,won2020scalable}. Prior work in this domain learns control policies that reproduce training motions, but the policies do not transfer to unseen test motions, nor do they estimate motion from video as our method does.

\section{Approach}

\begin{figure*}
    \centering
    \includegraphics[width=\textwidth]{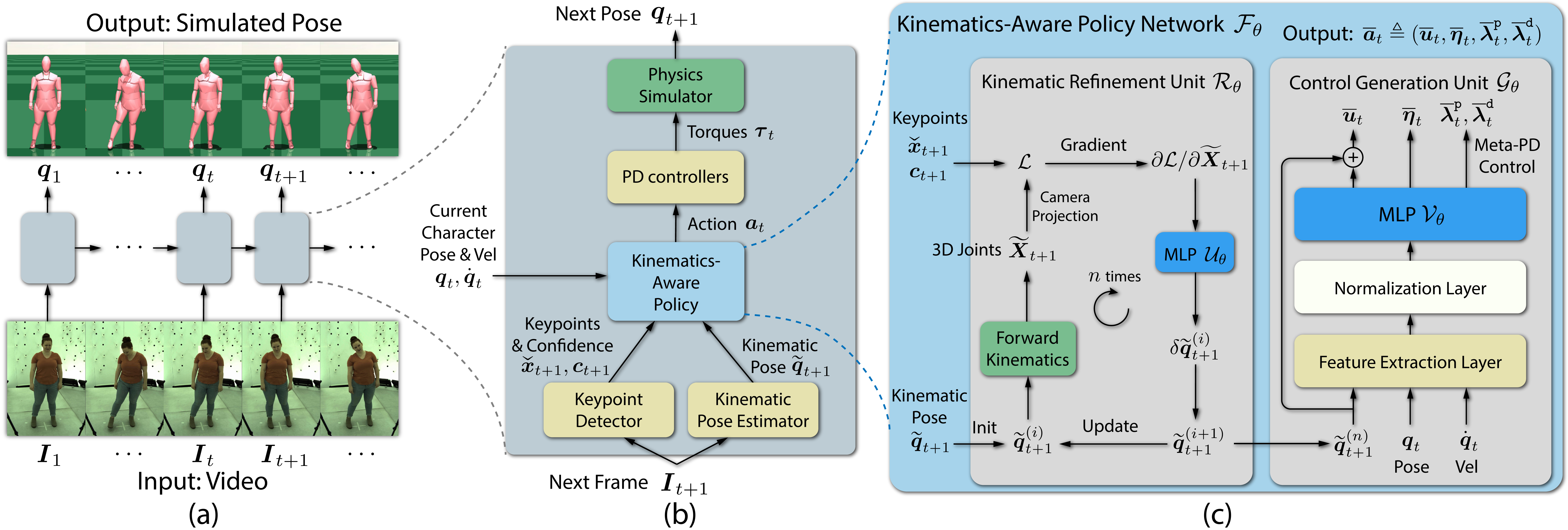}
    \caption{\textbf{Overview of our SimPoE framework.} (a) SimPoE is a physics-based causal temporal model. (b) At each frame (30Hz), the policy network $\mathcal{F}_\theta$ use the current pose $\bs{q}_t$, velocities $\dot{\bs{q}}_t$, and the next frame's estimated kinematic pose $\widetilde{\bs{q}}_{t+1}$ and keypoints $(\widecheck{\bs{x}}_{t+1}, \bs{c}_{t+1})$ to generate an action $\bs{a}_t$, which controls the character in the physics simulator (450Hz) via PD controllers to produce the next pose $\bs{q}_{t+1}$. (c) The policy network $\mathcal{F}_\theta$ outputs the mean action $\overline{\bs{a}}_t \triangleq (\overline{\bs{u}}_t, \overline{\bs{\eta}}_t,\overline{\bs{\lambda}}_t^\texttt{p},\overline{\bs{\lambda}}_t^\texttt{d})$. The kinematic refinement unit iteratively refines a kinematic pose estimate by learning pose updates. The refined pose $\widetilde{\bs{q}}^{(n)}_{t+1}$ is used by the control generation unit to produce the mean action $\overline{\bs{a}}_t$.
    \vspace{-3.5mm}}
    \label{fig:overview}
    \vspace{-2mm}
\end{figure*}

The overview of our SimPoE (\emph{\textbf{Sim}ulated Character Control for Human \textbf{Po}se \textbf{E}stimation}) framework is illustrated in Fig.~\ref{fig:overview}.
The input to SimPoE is a video $\bs{I}_{1:T} = (\bs{I}_1, \ldots, \bs{I}_T)$ of a person with $T$ frames. For each frame $\bs{I}_t$, we first use an off-the-shelf kinematic pose estimator to estimate an initial kinematic pose $\widetilde{\bs{q}}_t$, which consists of the joint angles and root translation of the person; we also extract 2D keypoints $\widecheck{\bs{x}}_t$ and their confidence $\bs{c}_t$ from $\bs{I}_t$ using a given pose detector (\emph{e.g.,} OpenPose~\cite{cao2017realtime})). As the estimated kinematic motion $\widetilde{\bs{q}}_{1:T} = (\widetilde{\bs{q}}_1, \ldots, \widetilde{\bs{q}}_T)$ is obtained without modeling human dynamics, it often contains physically-implausible poses with artifacts like jitter, foot sliding, and ground penetration. This motivates the main stage of our method, \emph{simulated character control}, where we model human dynamics with a proxy character inside a physics simulator. The character's initial pose $\bs{q}_1$ is set to $\widetilde{\bs{q}}_1$. At each time step $t$ shown in Fig.~\ref{fig:overview}\,(b), SimPoE learns a policy that takes as input the current character pose $\bs{q}_t$, velocities $\dot{\bs{q}}_t$, as well as the next frame's kinematic pose $\widetilde{\bs{q}}_{t+1}$ and keypoints $(\widecheck{\bs{x}}_{t+1},\bs{c}_{t+1})$ to produce an action that controls the character in the simulator to output the next pose $\bs{q}_{t+1}$. By repeating this causal process, we obtain the physically-grounded estimated motion $\bs{q}_{1:T} = (\bs{q}_1, \ldots, \bs{q}_T)$ of SimPoE.

\subsection{Automated Character Creation}
\label{sec:character_creation}
The character we use as a proxy to simulate human motion is created from skinned human mesh models, \emph{e.g.,} the SMPL model~\cite{loper2015smpl}, which can be recovered via SMPL-based pose estimation methods such as VIBE~\cite{kocabas2020vibe}. These skinned mesh models provide a skeleton of $B$ bones, a mesh of $V$ vertices, and a skinning weight matrix $\bs{W} \in \mathbb{R}^{V \times B}$ where each element $W_{ij}$ specifies the influence of the $j$-th bone's transformation on the $i$-th vertex's position. We can obtain a rigid vertex-to-bone association $\bs{A} \in \mathbb{R}^{V}$ by assigning each vertex~$i$ to the bone with the largest skinning weight for it: $A_i = \argmax_j W_{ij}$. With the vertex-to-bone association $\bs{A}$, we can then create the geometry of each bone by computing the 3D convex hull of all the vertices assigned to the bone. Assuming constant density, the mass of each bone is determined by the volume of its geometry. Our character creation process is fully automatic, is compatible with popular body mesh models (\emph{e.g.,} SMPL), and ensures proper body geometry and mass assignment.

\subsection{Simulated Character Control}

The task of controlling a character agent in physics simulation to generate desired human motions can be formulated as a Markov decision process (MDP), which is defined by a tuple $\mathcal{M} = (\mathcal{S}, \mathcal{A}, \mathcal{T}, R, \gamma)$ of states, actions, transition dynamics, a reward function, and a discount factor. The character agent interacts with the physics simulator according to a policy $\pi(\bs{a}_t|\bs{s}_t)$, which models the conditional distribution of choosing an action $\bs{a}_t \in \mathcal{A}$ given the current state $\bs{s}_t \in \mathcal{S}$ of the agent. Starting from some initial state $\bs{s}_1$, the character agent iteratively samples an action $\bs{a}_t$ from the policy $\pi$ and the simulation environment with transition dynamics $\mathcal{T}(\bs{s}_{t+1}|\bs{s}_t, \bs{a}_t)$ generates the next state $\bs{s}_{t+1}$ and gives the agent a reward $r_t$. The reward is assigned based on how well the character's motion aligns with the ground-truth human motion. The goal of our character control learning process is to learn an optimal policy $\pi^\ast$ that maximizes the expected return $J(\pi) = \mathbb{E}_{\pi}\left[\sum_{t}\gamma^t r_t\right]$ which translates to imitating the ground-truth motion as closely as possible. We apply a standard reinforcement learning algorithm (PPO~\cite{schulman2017proximal}) to solve for the optimal policy. In the following, we provide a detailed description of the states, actions and rewards of our control learning process. We then use a dedicated Sec.~\ref{sec:policy} to elaborate on our policy design.

\vspace{1mm}
\noindent\textbf{States.} The character state $\bs{s}_t \triangleq (\bs{q}_t, \dot{\bs{q}}_t, \widetilde{\bs{q}}_{t+1}, \widecheck{\bs{x}}_{t+1}, \bs{c}_{t+1})$ consists of the character's current pose $\bs{q}_t$, joint velocities (time derivative of the pose) $\dot{\bs{q}}_t$, as well as the estimated kinematic pose $\widetilde{\bs{q}}_{t+1}$, 2D keypoints $\widecheck{\bs{x}}_{t+1}$ and keypoint confidence $\bs{c}_{t+1}$ of the next frame. The state includes information of both the current frame ($\bs{q}_t,\dot{\bs{q}}_t$) and next frame ($\widetilde{\bs{q}}_{t+1}$, $\widecheck{\bs{x}}_{t+1}$,$\bs{c}_{t+1}$), so that the agent learns to take the right action $\bs{a}_{t}$ to transition from the current pose $\bs{q}_t$ to a desired next pose $\bs{q}_{t+1}$, \emph{i.e.,} pose close to the ground truth.

\vspace{1mm}
\noindent\textbf{Actions.} The policy $\pi(\bs{a}_t|\bs{s}_t)$ runs at 30Hz, the input video's frame rate, while our physics simulator runs at 450Hz to ensure stable simulation. This means one policy step corresponds to 15 simulation steps. One common design of the policy's action $\bs{a}_t$ is to directly output the torques $\bs{\tau}_t$ to be applied at each joint (except the root), which are used repeatedly by the simulator during the 15 simulation steps. However, finer control can be achieved by adjusting the torques at each step based on the state of the character. Thus, we follow prior work~\cite{peng2017learning,yuan2019ego} and use proportional-derivative (PD) controllers at each non-root joint to produce torques. With this design, the action $\bs{a}_t$ includes the target joint angles $\bs{u}_t$ of the PD controllers. At the $j$-th of the 15 simulation (PD controller) steps, the joint torques $\bs{\tau}_t$ are computed as 
\vspace{-2mm}
\begin{equation}
\label{eq:pd}
\bs{\tau}_t = \bs{k}_\texttt{p}\circ(\bs{u}_t - \bs{q}_t^\texttt{nr}) - \bs{k}_\texttt{d}\circ\dot{\bs{q}}_t^\texttt{nr},
\vspace{-1mm}
\end{equation}
where $\bs{k}_\texttt{p}$ and $\bs{k}_\texttt{d}$ are the parameters of the PD controllers, $\bs{q}_t^\texttt{nr}$ and $\dot{\bs{q}}_t^\texttt{nr}$ denote the joint angles and velocities of non-root joints at the start of the simulation step, and $\circ$ denotes element-wise multiplication. The PD controllers act like damped springs that drive joints to target angles $\bs{u}_t$, where $\bs{k}_\texttt{p}$ and $\bs{k}_\texttt{d}$ are the stiffness and damping of the springs. In Sec.~\ref{sec:meta_pd}, we will introduce a new control mechanism, meta-PD control, that allows $\bs{k}_\texttt{p}$ and $\bs{k}_\texttt{d}$ to be dynamically adjusted by the policy to achieve an even finer level of character control. With Meta-PD control, the action $\bs{a}_t$ includes elements $\bs{\lambda}_t^\texttt{p}$ and $\bs{\lambda}_t^\texttt{d}$ for adjusting $\bs{k}_\texttt{p}$ and $\bs{k}_\texttt{d}$ respectively. As observed in prior work~\cite{yuan2020residual}, allowing the policy to apply external residual forces to the root greatly improves the robustness of character control. Thus, we also add the residual forces and torques $\bs{\eta}_t$ of the root into the action $\bs{a}_t$. Overall, the action is defined as $\bs{a}_t \triangleq (\bs{u}_t, \bs{\eta}_t, \bs{\lambda}_t^\texttt{p}, \bs{\lambda}_t^\texttt{d})$.

\vspace{1mm}
\noindent\textbf{Rewards.} In order to learn the policy, we need to define a reward function that encourages the motion $\bs{q}_{1:T}$ generated by the policy to match the ground-truth motion $\widehat{\bs{q}}_{1:T}$. Note that we use $\;\widehat{\cdot}\;$ to denote ground-truth quantities. The reward $r_t$ at each time step is defined as the multiplication of four sub-rewards:
\vspace{-2mm}
\begin{align}
    r_t = r^\texttt{p}_t \cdot r^\texttt{v}_t \cdot r^\texttt{j}_t \cdot r^\texttt{k}_t\,.
\vspace{-1mm}
\end{align}
The pose reward $r^\texttt{p}_t$ measures the difference between the local joint orientations $\bs{o}^j_t$ and the ground truth~$\widehat{\bs{o}}^j_t$:
\vspace{-1.5mm}
\begin{equation}
\label{eq:r_p}
    r^\texttt{p}_t = \exp\left[-\alpha_\texttt{p}\left(\sum_{j=1}^J\|\bs{o}_t^j\ominus \widehat{\bs{o}}_t^j\|^2\right)\right],
\vspace{-1mm}
\end{equation}
where $J$ is the total number of joints, $\ominus$ denotes the relative rotation between two rotations, and $\|\cdot\|$ computes the rotation angle. The velocity reward $r^\texttt{v}_t$ measures the mismatch between joint velocities $\dot{\bs{q}}_t$ and the ground truth $\widehat{\dot{\bs{q}}}_t$:
\vspace{-1.5mm}
\begin{equation}
\label{eq:r_v}
    r^\texttt{v}_t = \exp\left[-\alpha_\texttt{v}\|\dot{\bs{q}}_t -  \widehat{\dot{\bs{q}}}_t\|^2\right].
\vspace{-1.5mm}
\end{equation}
The joint position reward $r^\texttt{j}_t$ encourages the 3D world joint positions $\bs{X}_t^j$ to match the ground truth  $\widehat{\bs{X}}_t^j$:
\vspace{-1.5mm}
\begin{equation}
\label{eq:r_j}
    r^\texttt{j}_t = \exp\left[-\alpha_\texttt{j}\left(\sum_{j=1}^J\|\bs{X}_t^j - \widehat{\bs{X}}_t^j\|^2\right)\right].
\vspace{-1.5mm}
\end{equation}
Finally, the keypoint reward $r^\texttt{k}_t$ pushes the 2D image projection $\bs{x}_t^j$ of the joints to match the ground truth $\widehat{\bs{x}}_t^j$:
\vspace{-1.5mm}
\begin{equation}
\label{eq:r_k}
    r^\texttt{k}_t = \exp\left[-\alpha_\texttt{k}\left(\sum_{j=1}^J\|\bs{x}_t^j - \widehat{\bs{x}}_t^j\|^2\right)\right].
\vspace{-1.5mm}
\end{equation}
Note that the orientations $\bs{o}^j_t$, 3D joint positions $\bs{X}_t^j$ and 2D image projections $\bs{x}_t^j$ are functions of the pose $\bs{q}_t$. The joint velocities $\dot{\bs{q}}_t$ are computed via finite difference. There are also weighting factors $\alpha_\texttt{p}, \alpha_\texttt{v}, \alpha_\texttt{j}, \alpha_\texttt{k}$ inside each reward.
These sub-rewards complement each other by matching different features of the generated motion to the ground-truth: joint angles, velocities, as well as 3D and 2D joint positions. Our reward design is multiplicative, which eases policy learning as noticed by prior work~\cite{won2020scalable}. The multiplication of the sub-rewards ensures that none of them can be overlooked in order to achieve a high reward.

\subsection{Kinematics-Aware Policy}
\label{sec:policy}

As the action $\bs{a}_t$ is continuous, we adopt a parametrized Gaussian policy $\pi_\theta(\bs{a}_t|\bs{s}_t) = \mathcal{N}(\overline{\bs{a}}_t, \bs{\Sigma})$ where the mean $\overline{\bs{a}}_t \triangleq (\overline{\bs{u}}_t, \overline{\bs{\eta}}_t, \overline{\bs{\lambda}}_t^\texttt{p}, \overline{\bs{\lambda}}_t^\texttt{d})$ is output by a neural network $\mathcal{F}_\theta$ with parameters $\theta$, and $\bs{\Sigma}$ is a fixed diagonal covariance matrix whose elements are treated as hyperparameters. The noise inside the Gaussian policy governed by $\bs{\Sigma}$ allows the agent to explore different actions around the mean action~$\overline{\bs{a}}_t$ and use these explorations to improve the policy during training. At test time, the noise is removed and the character agent always takes the mean action $\overline{\bs{a}}_t$ to improve performance. 

Now let us focus on the design of the policy network $\mathcal{F}_\theta$ that maps the state $\bs{s}_t$ to the mean action~$\overline{\bs{a}}_t$. Based on the design of $\bs{s}_t$, the mapping can be written as 
\vspace{-1mm}
\begin{equation}
    \overline{\bs{a}}_t = \mathcal{F}_\theta\left(\bs{q}_t, \dot{\bs{q}}_t, \widetilde{\bs{q}}_{t+1}, \widecheck{\bs{x}}_{t+1}, \bs{c}_{t+1}\right).
\vspace{-1mm}
\end{equation}
Recall that $\widetilde{\bs{q}}_{t+1}$ is the kinematic pose, $\widecheck{\bs{x}}_{t+1}$ and $\bs{c}_{t+1}$ are the detected 2D keypoints and their confidence, and that they are all information about the next frame. 
The overall architecture of our policy network $\mathcal{F}_\theta$ is illustrated in Fig.~\ref{fig:overview}\,(c).
The components $(\overline{\bs{u}}_t, \overline{\bs{\eta}}_t, \overline{\bs{\lambda}}_t^\texttt{p}, \overline{\bs{\lambda}}_t^\texttt{d})$ of the mean action $\overline{\bs{a}}_t$ are computed as follows:
\vspace{-1mm}
\begin{align}
\label{eq:ref}
    \widetilde{\bs{q}}_{t+1}^{(n)} &= \mathcal{R}_\theta\left(\widetilde{\bs{q}}_{t+1}, \widecheck{\bs{x}}_{t+1}, \bs{c}_{t+1}\right),\\
\label{eq:action}
    (\delta\overline{\bs{u}}_t, \overline{\bs{\eta}}_t, \overline{\bs{\lambda}}_t^\texttt{p}, \overline{\bs{\lambda}}_t^\texttt{d}) &=  \mathcal{G}_\theta\left(\widetilde{\bs{q}}_{t+1}^{(n)}, \bs{q}_t, \dot{\bs{q}}_t\right), \\
\label{eq:residual}
    \overline{\bs{u}}_t &= \widetilde{\bs{q}}_{t+1}^{(n)} + \delta\overline{\bs{u}}_t\,.
\vspace{-1mm}
\end{align}
In Eq.~\eqref{eq:ref}, $\mathcal{R}_\theta$ is a kinematic refinement unit that iteratively refines the kinematic pose $\widetilde{\bs{q}}_{t+1}$ using the 2D keypoints $\widecheck{\bs{x}}_{t+1}$ and confidence $\bs{c}_{t+1}$, and $\widetilde{\bs{q}}_{t+1}^{(n)}$ is the refined pose after $n$ iterations of refinement. Eq.~\eqref{eq:action} and \eqref{eq:residual} describe a control generation unit $\mathcal{G}_\theta$ that maps the refined pose $\widetilde{\bs{q}}_{t+1}^{(n)}$, current pose $\bs{q}_t$ and velocities $\dot{\bs{q}}_t$ to the components of the mean action $\overline{\bs{a}}_t$. Specifically, the control generation unit  $\mathcal{G}_\theta$ includes a hand-crafted feature extraction layer, a normalization layer (based on running estimates of mean and variance) and another MLP $\mathcal{V}_\theta$, as illustrated in Fig.~\ref{fig:overview}\,(c). As described in Eq.~\eqref{eq:residual}, an important design of $\mathcal{G}_\theta$  is a residual connection that produces the mean PD controller target angles $\overline{\bs{u}}_t$ using the refined kinematic pose~$\widetilde{\bs{q}}_{t+1}^{(n)}$,
where we ignore the root angles and positions in $\widetilde{\bs{q}}_{t+1}^{(n)}$ for ease of notation. This design builds in proper inductive bias since $\widetilde{\bs{q}}_{t+1}^{(n)}$ provides a good guess for the desired next pose $\bs{q}_{t+1}$ and thus a good base value for $\overline{\bs{u}}_t$. It is important to note that the PD controller target angles $\bs{u}_t$ do not translate to the same next pose $\bs{q}_{t+1}$ of the character, \emph{i.e.,} $\bs{q}_{t+1} \neq \bs{u}_t$. The reason is that the character is subject to gravity and contact forces, and under these external forces the joint angles $\bs{q}_{t+1}$ will not be $\bs{u}_t$ when the PD controllers reach their equilibrium. As an analogy, since PD controllers act like springs, a spring will reach a different equilibrium position when you apply external forces to it. Despite this, the next pose $\bs{q}_{t+1}$ generally will not be far away from $\bs{u}_t$ and learning the residual $\delta\overline{\bs{u}}_t$ to $\widetilde{\bs{q}}_{t+1}^{(n)}$ is easier than learning from scratch as we will demonstrate in the experiments. This design also synergizes the kinematics of the character with its dynamics as the kinematic pose $\widetilde{\bs{q}}_{t+1}^{(n)}$ is now tightly coupled with the input of the character's PD controllers that control the character in the physics simulator.

\vspace{2mm}
\noindent\textbf{Kinematic Refinement Unit.}
The kinematic refinement unit $\mathcal{R}_\theta$ is formed by an MLP $\mathcal{U}_\theta$ that maps a feature vector $\bs{z}$ (specific form will be described later) to a pose update:
\vspace{-1mm}
\begin{align}
    \delta \widetilde{\bs{q}}_{t+1}^{(i)} &= \mathcal{U}_\theta\left(\bs{z}\right),\\
    \widetilde{\bs{q}}_{t+1}^{(i+1)} &= \widetilde{\bs{q}}_{t+1}^{(i)} + \delta \widetilde{\bs{q}}_{t+1}^{(i)}\,,
\end{align}
where $i$ denotes the $i$-th refinement iteration and $\widetilde{\bs{q}}_{t+1}^{(0)} = \widetilde{\bs{q}}_{t+1}$.
To fully leverage the 2D keypoints and kinematic pose at hand, we design the feature $\bs{z}$ to be the gradient of the keypoint reprojection loss with respect to current 3D joint positions, inspired by recent work~\cite{song2020human} on kinematic body fitting. The purpose of using the gradient is not to minimize the reprojection loss, but to use it as an informative kinematic feature to learn a pose update that eventually results in stable and accurate control of the character; there is no explicit minimization of the reprojection loss in our formulation.
Specifically, we first obtain the 3D joint positions $\widetilde{\bs{X}}_{t+1} = \text{FK}(\widetilde{\bs{q}}_{t+1}^{(i)})$ through forward kinematics and then compute the reprojection loss as
\vspace{-2mm}
\begin{equation}
    \mathcal{L}(\widetilde{\bs{X}}_{t+1}) = \sum_{j=1}^J \left\| \Pi \left(\widetilde{\bs{X}}_{t+1}^j\right) - \widecheck{\bs{x}}_{t+1}^j\right\|^2\cdot c_{t+1}^j\,,
\vspace{-3mm}
\end{equation}
where $\widetilde{\bs{X}}_{t+1}^j$ denotes the $j$-th joint position in $\widetilde{\bs{X}}_{t+1}$, $\Pi(\cdot)$ denotes the perspective camera projection, and $(\widecheck{\bs{x}}_{t+1}^j, c_{t+1}^j)$ are the $j$-th detected keypoint and its confidence. The gradient feature $\bs{z}\triangleq \partial \mathcal{L}/ \partial \widetilde{\bs{X}}_{t+1}$ is informative about the kinematic pose $\widetilde{\bs{q}}_{t+1}^{(i)}$ as it tells us how each joint should move to match the 2D keypoints $\widecheck{\bs{x}}_{t+1}^j$. It also accounts for keypoint uncertainty by weighting the loss with the keypoint confidence $c_{t+1}^j$. Note that $\bs{z}$ is converted to the character's root coordinate to be invariant of the character's orientation. The refinement unit integrates kinematics and dynamics as it utilizes a kinematics-based feature $\bs{z}$ to learn the update of a kinematic pose, which is used to produce dynamics-based control of the character. The joint learning of the kinematic refinement unit $\mathcal{R}_\theta$ and the control generation unit $\mathcal{G}_\theta$ ensures accurate and physically-plausible pose estimation.

\vspace{2mm}
\noindent\textbf{Feature Extraction Layer.} After refinement, the control generation unit $\mathcal{G}_\theta$ needs to extract informative features from its input to output an action that advances the character from the current pose $\bs{q}_t$ to the next pose $\bs{q}_{t+1}$. To this end, the feature extraction layer uses information from both the current frame and next frame. Specifically, the extracted feature includes $\bs{q}_t$, $\dot{\bs{q}}_t$, the current 3D joint positions $\bs{X}_t$, the pose difference vector between $\bs{q}_t$ and the refined kinematic pose $\widetilde{\bs{q}}_{t+1}^{(n)}$, and the difference vector between $\bs{X}_t$ and the next-frame joint position $\widetilde{\bs{X}}_{t+1}$ computed from $\widetilde{\bs{q}}_{t+1}^{(n)}$. All features are converted to the character's root coordinate to be orientation-invariant and encourage robustness against variations in absolute pose encountered at test time.

\vspace{-0.5mm}
\subsection{Meta-PD control}
\vspace{-0.5mm}
\label{sec:meta_pd}
PD controllers are essential in our approach as they relate the kinematics and dynamics of the character by converting target joint angles in pose space to joint torques. However, an undesirable aspect of PD controllers is the need to specify the parameters $\bs{k}_\texttt{p}$ and $\bs{k}_\texttt{d}$ for computing the joint torques $\bs{\tau}_t$ as described in Eq.~\eqref{eq:pd}. It is undesirable because (i) manual parameter tuning requires significant domain knowledge and (ii) even carefully designed parameters can be suboptimal. The difficulty, here, lies in balancing the ratio between $\bs{k}_\texttt{p}$ and $\bs{k}_\texttt{d}$. Large ratios can lead to unstable and jittery motion while small values can result in motion that is too smooth and lags behind ground truth.

Motivated by this problem, we propose meta-PD control, a method that allows the policy to dynamically adjust $\bs{k}_\texttt{p}$ and $\bs{k}_\texttt{d}$ based on the state of the character. Specifically, given some initial values $\bs{k}'_\texttt{p}$ and $\bs{k}'_\texttt{d}$, the policy outputs $\lambda_\texttt{p}$ and $\lambda_\texttt{d}$ as additional elements of the action $\bs{a}_t$ that act to scale $\bs{k}'_\texttt{p}$ and $\bs{k}'_\texttt{d}$. Moreover, we take this idea one step further and let the policy output two sequences of scales $\bs{\lambda}_t^\texttt{p}=(\lambda_{t1}^\texttt{p}, \ldots, \lambda_{tm}^\texttt{p})$ and $\bs{\lambda}_t^\texttt{d}=(\lambda_{t1}^\texttt{d}, \ldots, \lambda_{tm}^\texttt{d})$ where $m=15$ corresponds to the number of PD controller (simulation) steps during a policy step. The PD controller parameters $\bs{k}_\texttt{p}$ and $\bs{k}_\texttt{d}$ at the $j$-th step of the 15 PD controller steps are then computed as follows:
\vspace{-1mm}
\begin{equation}
    \bs{k}_\texttt{p} = \lambda_{tj}^\texttt{p}\bs{k}'_\texttt{p}, \quad \bs{k}_\texttt{d} = \lambda_{tj}^\texttt{d}\bs{k}'_\texttt{d}\,.
\vspace{-1mm}
\end{equation}
Instead of using fixed $\bs{k}_\texttt{p}$ and $\bs{k}_\texttt{d}$, meta-PD control allows the policy to plan the scaling of $\bs{k}_\texttt{p}$ and $\bs{k}_\texttt{d}$ through the 15 PD controller steps to have more granular control over the torques produced by the PD controllers, which in turn enables a finer level of character control. With meta-PD control, the action $\bs{a}_t$ is now defined as $\bs{a}_t \triangleq (\bs{u}_t, \bs{\eta}_t, \bs{\lambda}_t^\texttt{p}, \bs{\lambda}_t^\texttt{d})$.

\vspace{-1mm}
\section{Experiments}

\begin{figure*}[h!]
    \centering
    \includegraphics[width=\textwidth]{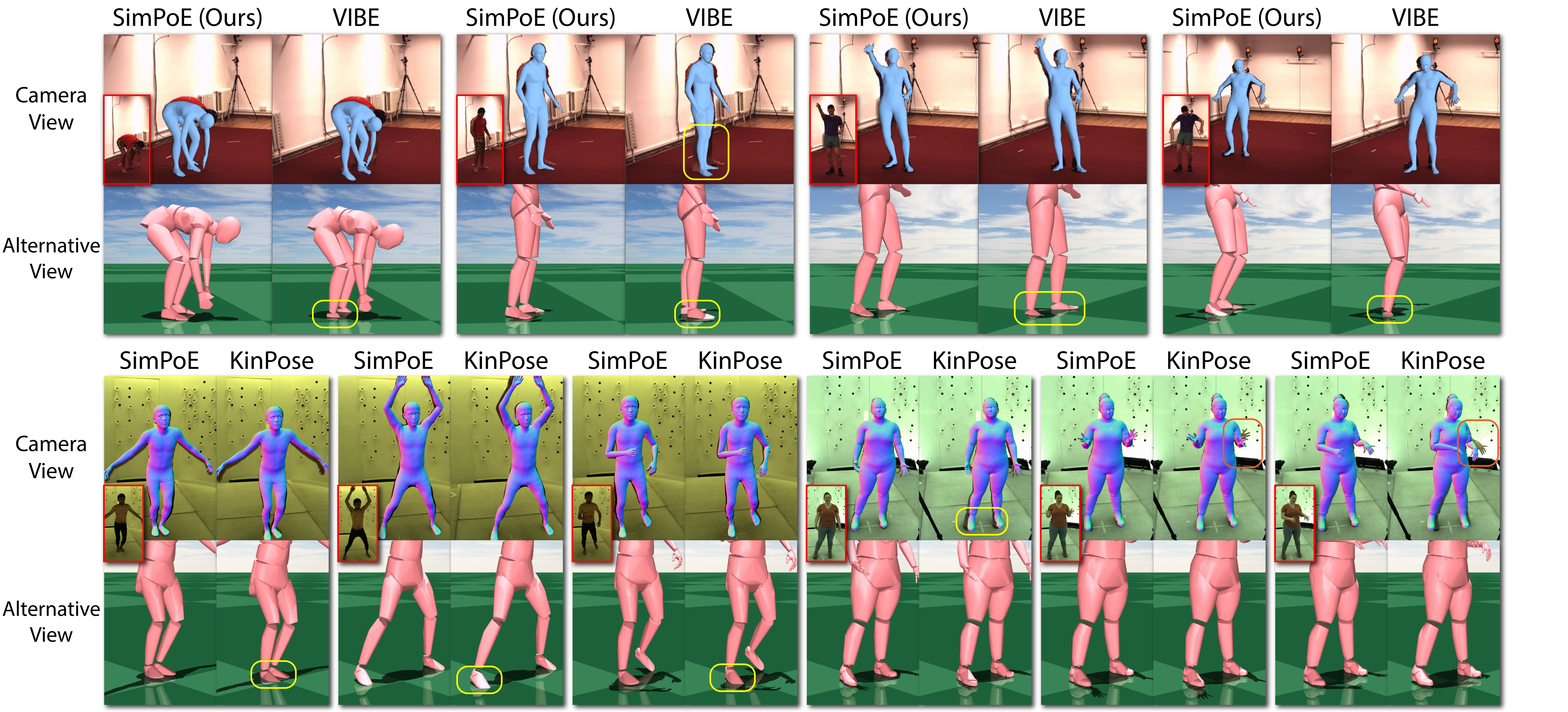}
    \caption{\textbf{Visualization} of estimated poses in the camera view and an alternative view. SimPoE estimates more accurate poses and foot contact. Pose mismatch and ground penetration are highlighted with boxes. Please see the \href{https://www.ye-yuan.com/simpoe}{supplementary video} for more comparisons. \vspace{-3.5mm}}
    \label{fig:qual_res}
    \vspace{-2.5mm}
\end{figure*}

\noindent\textbf{Datasets.}
We perform experiments on two large-scale human motion datasets. The first dataset is Human3.6M~\cite{ionescu2013human3}, which includes 7 annotated subjects captured at 50Hz and a total of 1.5 million training images. Following prior work~\cite{kolotouros2019learning,kocabas2020vibe,moon2020i2l}, we train our model on 5 subjects (S1, S5, S6, S7, S8) and test on the other 2 subjects (S9, S11). We subsample the dataset to 25Hz for both training and testing.
The second dataset we use is an in-house human motion dataset that also contains \emph{detailed finger motion}. It consists of 3 subjects captured at 30Hz performing various actions from free body motions to natural conversations. There are around 335k training frames and 87k test frames. Our in-house dataset has complex skeletons with twice more joints than the SMPL model, including fingers. The body shape variation among subjects is also greater than that of SMPL, which further evaluates the robustness of our approach.

\vspace{2mm}
\noindent\textbf{Metrics.} We use both pose-based and \emph{physics-based} metrics for evaluation. To assess pose accuracy, we report mean per joint position error (MPJPE) and Procrustes-aligned mean per joint
position error (PA-MPJPE). We also use three physics-based metrics that measure jitter, foot sliding, and ground penetration, respectively. For jitter, we compute the difference in acceleration (Accel) between the predicted 3D joint and the ground-truth. For foot sliding (FS), we find body mesh vertices that contact the ground in two adjacent frames and compute their average displacement within the frames. For ground penetration (GP), we compute the average distance to the ground for mesh vertices below the ground. The units for these metrics are millimeters (mm) except for Accel (mm/frame$^2$). MPJPE, PA-MPJPE and Accel are computed in the root-centered coordinate.

\subsection{Implementation Details.}
 \noindent\textbf{Character Models.}
 We use MuJoCo~\cite{todorov2012mujoco} as the physics simulator. For the character creation process in Sec.~\ref{sec:character_creation}, we use VIBE~\cite{kocabas2020vibe} to recover an SMPL model for each subject in Human3.6M. Each MuJoCo character created from the SMPL model has 25 bones and 76 degrees of freedom (DoFs). For our in-house motion dataset, we use non-rigid ICP~\cite{amberg2007optimal} and linear blend skinning~\cite{kavan2007skinning} to reconstruct a skinned human mesh model for each subject. Each of these models has fingers and includes 63 bones and 114 DoFs.

\input{tables/table_quan}

\begin{figure*}[ht!]
\floatsetup{captionskip=5pt}
\begin{floatrow}
\hspace{-4mm}
\capbtabbox{%
  \input{tables/table_abl}
}{%
  \caption{Ablation studies on Human3.6M and our in-house motion dataset.}
  \label{table:abl}
}
\hspace{-1cm}
\ffigbox[\FBwidth]{%
     \centering
    \includegraphics[trim=0cm 2mm 0cm 3mm, clip=true, width=0.29\textwidth]{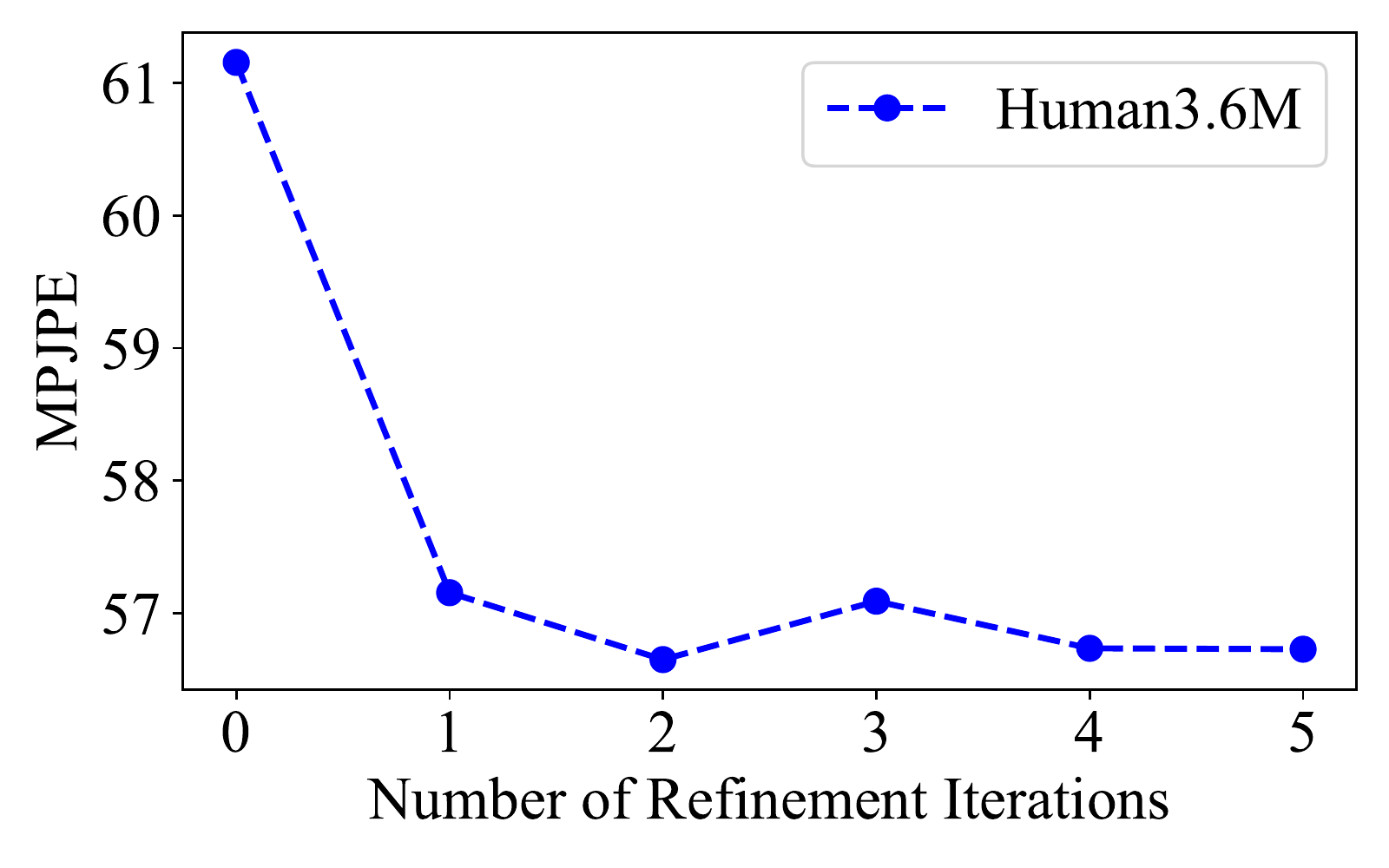}
    \vspace{-1mm}
}{%
  \caption{Effect of refinement unit.}
  \label{fig:plot}
}
\vspace{-3mm}
\end{floatrow}
\end{figure*}

\vspace{2mm}
\noindent\textbf{Initialization.}
For Human3.6M, we use VIBE to provide the initial kinematic motion $\widetilde{\bs{q}}_{1:T}$. For our in-house motion dataset, since our skinned human models have more complex skeletons and meshes than the SMPL model, we develop our own kinematic pose estimator, which is detailed in Appendix~\ref{sec:app_data}. To recover the global root position of the person, we assume the camera intrinsic parameters are calibrated and optimize the root position by minimizing the reprojection loss of 2D keypoints, similar to the kinematic initialization in~\cite{shimada2020physcap}.

\vspace{2mm}
\noindent\textbf{Other Details.} The kinematic refinement unit in the policy network refines the kinematic pose $n=5$ times. To facilitate learning, we first pretrain the refinement unit with supervised learning using an MSE loss on the refined kinematic pose. The normalization layer in the policy computes the running average of the mean and variance of the input feature during training, and uses it to produce a normalized feature. Our learned policy runs at 38 FPS on a standard PC with an Intel Core i9 Processor. More implementation details such as training procedures and hyperparameter settings can be found in Appedix~\ref{sec:app_detail}.

\subsection{Comparison to state-of-the-art methods}
We compare SimPoE against state-of-the-art monocular 3D human pose estimation methods, including both kinematics-based (VIBE~\cite{kocabas2020vibe}, NeurGD~\cite{song2020human}) and physics-based (PhysCap~\cite{shimada2020physcap}, EgoPose~\cite{yuan2019ego}) approaches. The results of VIBE and EgoPose are obtained using their publicly released code and models. As PhysCap and NeurGD have not released their code, we directly use the reported results on Human3.6M from the PhysCap paper and implement our own version of NeurGD. Table~\ref{table:quan} summarizes the quantitative results on Human3.6M and the in-house motion dataset. On Human3.6M, we can observe that our method, SimPoE, outperforms previous methods in pose accuracy as indicated by the smaller MPJPE and PA-MPJPE. In particular, SimPoE shows large pose accuracy improvements over state-of-the-art physics-based approaches (EgoPose~\cite{yuan2019ego} and PhysCap~\cite{shimada2020physcap}), reducing the MPJPE almost by half. For physics-based metrics (Accel, FS and GP), SimPoE also outperforms prior methods by large margins. It means that SimPoE significantly reduces the physical artifacts -- jitter (Accel), foot sliding (FS), and ground penetration (GP), which particularly deteriorate the results of kinematic methods (VIBE~\cite{kocabas2020vibe} and NeurGD~\cite{song2020human}). On the in-house motion dataset, SimPoE again outperforms previous methods in terms of both pose-based and physics-based metrics. In the table, KinPose denotes our own kinematic pose estimator used by SimPoE. We note that the large acceleration error (Accel) of EgoPose is due to the frequent falling of the character, which is a common problem in physics-based methods since the character can lose balance when performing agile motions. The learned policy of SimPoE is robust enough to stably control the character without falling, which prevents irregular accelerations.

We also provide qualitative comparisons in Fig.~\ref{fig:qual_res}, where we show the estimated poses in the camera view and the same poses rendered from an alternative view. The alternative view shows that SimPoE can estimate foot contact with the ground more accurately and without penetration. As the quality and physical plausibility of the estimated motions are best seen in videos, please refer to the \href{https://www.ye-yuan.com/simpoe}{supplementary video} for additional qualitative results and comparisons.

\subsection{Ablation Studies}

To further validate our proposed approach, we conduct extensive ablation studies to investigate the contribution of each proposed component to the performance. Table~\ref{table:abl} summarizes the results where we train different variants of SimPoE by removing a single component each time. First, we can observe that both meta-PD control and the kinematic refinement unit contribute to better pose accuracy as indicated by the corresponding ablations (w/o Meta-PD and w/o Refine). Second, the ablation (w/o ResAngle) shows that it is important to have the residual connection in the policy network for producing the mean PD controller target angles $\overline{\bs{u}}_t$. Next, the residual forces $\bs{\eta}_t$ we use in action~$\bs{a}_t$ are also indispensable as demonstrated by the drop in performance of the variant (w/o ResForce). Without the residual forces, the policy is not robust and the character often falls down as indicated by the large acceleration error (Accel). Finally, it is evident from the ablation (w/o FeatLayer) that our feature extraction layer in the policy is also instrumental, because it extracts informative features of both the current frame and next frame to learn control that advances the character to the next pose. We also perform ablations to investigate how the number of refinement iterations in the policy affects pose accuracy. As shown in Fig.~\ref{fig:plot}, the performance gain saturates around 5 refinement iterations.

\vspace{-2mm}
\section{Discussion and Future Work}
\vspace{-1mm}
In this work, we demonstrate that modeling both kinematics and dynamics improves the accuracy and physical plausibility of 3D human pose estimation from monocular video. Our approach, \mbox{SimPoE}, unifies kinematics and dynamics by integrating image-based kinematic inference and physics-based character control into a joint reinforcement learning framework. It runs in real-time, is compatible with advanced physics simulators, and addresses several drawbacks of prior physics-based approaches.

However, due to its physics-based formulation, \mbox{SimPoE} depends on 3D scene modeling to enforce contact constraints during motion estimation. This hinders direct evaluation on in-the-wild datasets, such as 3DPW~\cite{von2018recovering}, which includes motions such as climbing stairs or even trees. Future work may include integration of video-based 3D scene reconstruction to address this limitation. 

{\small
\bibliographystyle{ieee_fullname}
\bibliography{ref}
}

\clearpage
\onecolumn
\appendix
\section{In-House Motion Dataset and Kinematic Pose Estimator}
\label{sec:app_data}
\noindent\textbf{In-House Dataset.}
Our in-house motion dataset uses a more complex skeleton model (with twice as many joints, including fingers) than SMPL~\cite{loper2015smpl}. To recover the human skinning mesh model of each subject, we use an offline process that uses multiview 3D reconstruction to produce a person-specific skinning template with 32 bone scaling parameters based on non-rigid ICP deformation~\cite{amberg2007optimal} and linear blend skinning~\cite{kavan2007skinning}. For motion, we solve for 94 local joint angles (degrees of freedom) and the global 3D position of 159 joints, including finger joints, for every frame.

\vspace{2mm}
\noindent\textbf{Kinematic Pose Estimator.}
Since existing kinematic pose estimators, such as VIBE~\cite{kocabas2020vibe}, cannot be directly applied to the in-house dataset due to the dataset's more complex skeletons and skinning models, we design a simple kinematic tracker (``KinPose'' in the main paper) that also uses monocular inputs to produce kinematic pose estimates. The model does not have any temporal component, outputting both 2D keypoint heatmaps and joint angles frame-by-frame. These two outputs are required by our approach (SimPoE) and NeurGD~\cite{sun2019human}.
Below, we detail the network architecture and training procedure of this model, which are typical for such models as the performance of SimPoE is not sensitive to these design choices.

\vspace{2mm}
\noindent\textbf{Network Architecture.}
 We use a 3-stage cascaded network~\cite{Wei2016convolutional} with a backbone based on ResNet-50~\cite{He2015deep}. The output of the network at each stage is a tensor of $m+n$ channels with a spatial size that is $8$x smaller than the input image, where $m = 77$ is the number of heatmap channels for 2D keypoints (a subset of the $159$ joints), and $n = 94+6$ is the number of local joint angles and global pose dimensions. Each of the $94$ angle channels is an ``angle map'' corresponding to a joint. The final output angle (scalar) is calculated by summing the element-wise product between an angle map and its corresponding keypoint heatmaps, where the correspondence is defined based on the skeleton. This design is a type of attention mechanism, which encourages the model to predict the angle of a joint based only on relevant image regions.

\vspace{2mm}
\noindent\textbf{Training.}
At each stage of the network, we apply $L_2$ losses on heatmaps, 2D keypoints, 3D joint positions, and joint angles to train the model, similar to VIBE~\cite{kocabas2020vibe}.

\section{Additional Implementation Details}
\label{sec:app_detail}

\begin{table}[h]
\footnotesize
\centering
\begin{tabular}{@{\hskip 0.5cm}l@{\hskip 2cm}c@{\hskip 0.5cm}}
\toprule
Parameter & Value\\ \midrule
Num. of time steps & 50000 \\
Num. of epochs & 2000 \\
Num. of policy updates per epoch & 10 \\
Policy step size & $5\times 10^{-5}$\\
Value step size & $3\times 10^{-4}$\\
PPO clip $\epsilon$ & 0.2 \\
Discount factor $\gamma$ & 0.95 \\
GAE coefficient $\lambda$ & 0.95 \\
Reward weights ($\alpha_\texttt{p}, \alpha_\texttt{v}, \alpha_\texttt{j}, \alpha_\texttt{k}$) (Human3.6M) & (30, 0.2, 100, 0.02) \\
Reward weights ($\alpha_\texttt{p}, \alpha_\texttt{v}, \alpha_\texttt{j}, \alpha_\texttt{k}$) (In-house) & (60, 0.2, 300, 0.02) \\
Elements of diagonal covariance $\bs{\Sigma}$ (Human3.6M) & 0.1 \\
Elements of diagonal covariance $\bs{\Sigma}$ (In-house) & 0.05 \\
Residual force scale & 500 \\
\bottomrule
\end{tabular}
\vspace{1mm}
\caption{Hyperparameters for experiments on Human3.6M~\cite{ionescu2013human3} and our in-house motion dataset.}
\label{table:hyper}
\end{table}

For both Human3.6M and our in-house motion dataset, our method uses the same hyperparameter settings unless stated otherwise. Table~\ref{table:hyper} summarizes the hyperparameter setting.

\vspace{2mm}
\noindent\textbf{Policy Network.}
The learnable parts in the policy network are the two MLPs, \emph{i.e.,} $\mathcal{U}_\theta$ inside the kinematic refinement unit and $\mathcal{V}_\theta$ inside the control generation unit. We use ReLU activations for both $\mathcal{U}_\theta$ and $\mathcal{V}_\theta$. The MLP $\mathcal{U}_\theta$ consists of hidden layers with size (256, 512, 256). The MLP $\mathcal{V}_\theta$ contains hidden layers with size (2048, 1024).

\vspace{2mm}
\noindent\textbf{Policy Training.} For Human3.6M, we train a single policy using data from all the training subjects and directly transfer the policy to test subjects, so it is a cross-subject experiment. For our in-house motion dataset, due to the large variation of body proportion and shape, we train a model for each subject using subject-specific data and test on separate data. All baselines are trained using the same data as our method. For learning the policy, each RL episode is constructed by randomly sampling a video segment of 200 frames from all training data. For the initial pose $\bs{q}_1$ of the character, we initialize it to the refined kinematic pose $\widetilde{\bs{q}}_1^{(n)}$. For the initial velocity $\dot{\bs{q}}_1$, we set it to the kinematic velocity $\widetilde{\dot{\bs{q}}}_1^{(n)}$ computed using finite differences. The episode is terminated when the end frame is reached or the character's root height is 0.5 below the root height of the kinematic pose (i.e., to detect if the character has lost balance). We train the policy $\pi_\theta$ for 2000 epochs. For each epoch, we keep collecting data by sampling RL episodes until the total number of time steps reaches 50000. The reward weighting factors ($\alpha_\texttt{p}, \alpha_\texttt{v}, \alpha_\texttt{j}, \alpha_\texttt{k}$) are set to (30, 0.2, 100, 0.02) for Human3.6M and (60, 0.2, 300, 0.02) for the in-house dataset. For Human3.6M, we only have access to ground-truth 3D joint positions but not ground-truth joint angles, so we use the refined kinematic pose as pseudo-ground truth (for regularization) when computing rewards that need ground-truth joint angles. The elements of the policy's diagonal covariance matrix $\bs{\Sigma}$ are set to 0.1 for Human3.6M and 0.05 for our in-house motion dataset. The residual forces $\bs{\eta}_t$ output by the policy is scaled by 500 before being input to the physics simulator. We use the proximal policy optimization (PPO~\cite{schulman2017proximal}) to learn the policy $\pi_\theta$. The discount factor for the Markov decision process (MDP) is 0.95. We use the generalized advantage estimator GAE($\lambda$)~\cite{schulman2015high} to compute the advantage estimate for policy gradient and the GAE coefficient $\lambda$ is 0.95. At each epoch, the policy is updated 10 times using Adam~\cite{kingma2014adam} with a step size $5\times 10^{-5}$. The clipping coefficient $\epsilon$ in PPO is set to 0.2. Since PPO is an actor-critic based method, it also learns a value function that mirrors the design of the policy but outputs a single value estimate. The value function is updated using Adam with a step size $3\times 10^{-4}$ whenever the policy is updated.

\end{document}

%% file: tables/table_quan.tex
\setlength{\tabcolsep}{3pt}
\begin{table}[t]
\vspace{2mm}
\footnotesize
\centering
\resizebox{\linewidth}{!}{
\begin{tabular}{@{\hskip 1mm}lcccccc@{\hskip 1mm}}
\toprule
\multicolumn{7}{c}{Human3.6M}\\
\midrule
Method & Physics  & MPJPE $\downarrow$ & PA-MPJPE $\downarrow$ & Accel $\downarrow$ & FS $\downarrow$ & GP $\downarrow$ \\ \midrule
VIBE~\cite{kocabas2020vibe} & \xmark & 61.3	& 43.1 & 15.2 & 15.1 & 12.6\\
NeurGD*~\cite{song2020human} & \xmark & 57.3 & 42.2 & 14.2 & 16.7 & 24.4\\
PhysCap~\cite{shimada2020physcap} & \cmark & 113.0 & 68.9 & - & - & - \\
EgoPose~\cite{yuan2019ego} & \cmark & 130.3 & 79.2 & 31.3 & 5.9  & 3.5 \\
SimPoE (Ours) & \cmark & \textbf{56.7} & \textbf{41.6} & \textbf{6.7} & \textbf{3.4} & \textbf{1.6}\setcounter{rownum}{0}\\
\midrule
\multicolumn{7}{c}{In-House Motion Dataset}\\
\midrule
Method & Physics  & MPJPE $\downarrow$ & PA-MPJPE $\downarrow$ & Accel $\downarrow$ & FS $\downarrow$ & GP $\downarrow$ \\ \midrule
KinPose  & \xmark & 49.7  & 40.4  & 12.8 & 6.4 & 3.9 \\
NeurGD*~\cite{song2020human} & \xmark & 36.7  & 30.9  & 16.2 & 7.7 & 3.6 \\
EgoPose~\cite{yuan2019ego} & \cmark & 202.2 & 131.4 & 32.6 & 2.2 & 0.5 \\
SimPoE (Ours) & \cmark & \textbf{26.6} & \textbf{21.2} & \textbf{8.4} & \textbf{0.5} & \textbf{0.1}\\
\bottomrule
\end{tabular}
}
\vspace{1pt}
\caption{Results of pose-based (MPJPE, PA-MPJPE) and physics-based (Accel, FS, GP) metrics on Human3.6M and our in-house motion dataset. Symbol ``-'' means results are not available and ``*'' means self-implementation (better results than the original paper). \vspace{-6.2mm}}
\label{table:quan}
\vspace{-3mm}
\end{table}

%% file: tables/table_abl.tex
\setlength{\tabcolsep}{3pt}
\resizebox{0.68\textwidth}{!}{
\begin{tabular}{@{\hskip 1mm}lcccccccccccc@{\hskip 1mm}}
\toprule
\multirow{3}{*}[2pt]{Method} & \multicolumn{5}{c}{Human3.6M} & \multicolumn{5}{c}{In-House Motion Dataset} \\
\cmidrule(l{0mm}r{1mm}){2-6} \cmidrule(l{0.5mm}r{0mm}){7-11}
 & MPJPE $\downarrow$ & PA-MPJPE $\downarrow$ & Accel $\downarrow$ & FS $\downarrow$ & GP $\downarrow$ & MPJPE $\downarrow$ & PA-MPJPE $\downarrow$ & Accel $\downarrow$ & FS $\downarrow$ & GP $\downarrow$ \\ \midrule
w/o Meta-PD & 59.9 & 44.7 & \textbf{5.9} & \textbf{2.2} & \textbf{1.4} & 39.8  & 31.7  & 7.1  & \textbf{0.4} & \textbf{0.1} \\
w/o Refine & 61.2 & 43.5 & 8.0 & 3.4 & 2.0 & 47.9  & 38.9  & 9.6  & 0.6 & \textbf{0.1} \\
w/o ResAngle & 68.7 & 51.0 & 6.4 & 4.1 & 2.1 & 193.4 & 147.6 & \textbf{6.5}  & 0.9 & 0.3 \\
w/o ResForce & 115.2 & 65.1 & 23.5 & 6.1 & 3.2 & 48.4  & 31.3  & 12.5 & 0.9 & 0.3\\
w/o FeatLayer & 81.4 & 47.6	& 9.3 & 5.0	& 1.8 & 36.9  & 27.5  & 9.5  & 0.6 & \textbf{0.1} \\
SimPoE (Ours) & \textbf{56.7} & \textbf{41.6} & 6.7 & 3.4 & 1.6 & \textbf{26.6} & \textbf{21.2} & 8.4 & 0.5 & \textbf{0.1}\\
\bottomrule
\end{tabular}
}